\documentclass[a4paper,twoside]{article}

\pdfoutput=1

\usepackage[inline]{enumitem}
\usepackage{hyperref}
\usepackage{tabularx}
\usepackage{xfrac}
\usepackage{numprint}
\usepackage{float}
\usepackage[T1]{fontenc}
\usepackage{stfloats}
\usepackage{acronym}
\usepackage{booktabs}
\usepackage{censor}

\usepackage{epsfig}
\usepackage{subcaption}
\usepackage{calc}
\usepackage{amssymb}
\usepackage{amstext}
\usepackage{amsmath}
\usepackage{amsthm}
\usepackage{multicol}
\usepackage{pslatex}
\usepackage{apalike}
\usepackage[bottom]{footmisc}
\usepackage{SCITEPRESS}     

\newacro{AutoML}[AutoML]{automated machine learning}
\newacro{CRISP-DM}[CRISP-DM]{Cross Industry Standard Process for Data Mining}
\newacro{ML}[ML]{machine learning}
\newacro{SME}[SME]{small and medium-sized enterprise}
\newacro{POC}[POC]{proof of concept}
\newacro{MVP}[MVP]{minimal viable product}
\newacro{MLP}[MLP]{multi-layer perceptron}
\newacro{MAE}[MAE]{mean absolute error}
\newacro{RMSE}[RMSE]{root mean squared error}
\newacro{MAPE}[MAPE]{mean absolute percentage error}
\newacro{SVR}[SVR]{support vector regression}
\newacro{kNN}[kNN]{\(k\)-nearest neighbors}
\newacro{HPO}[HPO]{hyperparameter optimization}
\newacro{RF}[RF]{random forest}
\newacro{PQ}[PQ]{Prediction Quality}
\newacro{KL}[KL]{Knowledge Levels}
\newacro{TR}[TR]{Training Runtime}
\newacro{MES}[MES]{method evaluation score}
\newacro{LOC}[LOC]{lines of code}
\newacro{NNI}[NNI]{Neural Network Intelligence}
\newacro{XAI}[XAI]{explainable artificial intelligence}

\newcommand*{\name}[1]{#1}
\newcommand{\res}[2]{#1\(\,\pm\,\)#2}
\newcommand{\bres}[2]{\textbf{#1}\(\,\pm\,\)\textbf{#2}}

\newcommand{\eg}{e.g.,~}

\begin{document}

\title{Benchmarking Automated Machine Learning Methods\\ for Price Forecasting Applications}

\author{
\authorname{
Horst St\"uhler\sup{1}\orcidAuthor{0000-0002-7638-1861},
Marc-Andr\'e Z\"oller\sup{2}\orcidAuthor{0000-0001-8705-9862},
Dennis Klau\sup{3}\orcidAuthor{0000-0003-3618-7359},
Alexandre Beiderwellen-Bedrikow\sup{1}\orcidAuthor{0000-0001-7934-8410} and
Christian Tutschku\sup{3}\orcidAuthor{0000-0003-0401-5333}
}
\affiliation{\sup{1}Zeppelin GmbH, Graf-Zeppelin-Platz 1, 85766 Garching, Germany}
\affiliation{\sup{2}USU Software AG, R\"uppurrer Str. 1, 76137 Karlsruhe, Germany}
\affiliation{\sup{3}Fraunhofer IAO, Nobelstraße 12, 70569 Stuttgart, Germany}
\email{\{horst.stuehler, alexandre.bedrikow\}@zeppelin.com, marc.zoeller@usu.com, \\ \{dennis.klau, christian.tutschku\}@iao.fraunhofer.de}
}



\abstract{
Price forecasting for used construction equipment is a challenging task due to spatial and temporal price fluctuations. It is thus of high interest to automate the forecasting process based on current market data. Even though applying \ac{ML} to these data represents a promising approach to predict the residual value of certain tools, it is hard to implement for \aclp{SME} due to their insufficient \ac{ML} expertise. To this end, we demonstrate the possibility of substituting manually created \ac{ML} pipelines with \ac{AutoML} solutions, which automatically generate the underlying pipelines. We combine \ac{AutoML} methods with the domain knowledge of the companies. Based on the \acs{CRISP-DM} process, we split the manual \ac{ML} pipeline into a \acl{ML} and non-\acl{ML} part. To take all complex industrial requirements into account and to demonstrate the applicability of our new approach, we designed a novel metric named \acl{MES}, which incorporates the most important technical and non-technical metrics for quality and usability. Based on this metric, we show in a case study for the industrial use case of price forecasting, that domain knowledge combined with \ac{AutoML} can weaken the dependence on \ac{ML} experts for innovative \aclp{SME} which are interested in conducting such solutions.
}

\onecolumn \maketitle \normalsize \setcounter{footnote}{0} \vfill

\section{\uppercase{Introduction}}
\label{sec:introduction}

Price forecasting is crucial for companies dealing with used assets whose price depends on availability and demand varying spatially and over time. Especially the sector of heavy construction equipment dealers and rental companies relies heavily on accurate price predictions. Determining the current and future residual value of their fleet allows construction equipment dealers to identify the optimal time to resell individual pieces of machinery \cite{residualvalue_lucko2007,chiteri2018cash}. Although several data-driven methods have been proposed to forecast the heavy equipment's residual value \cite{lucko2003statistical,lucko2004predicting,fan2008assessing,lucko2011modeling,zong2017maintenance,milovsevic2020determination}, price forecasting in practice is still mainly performed manually due to the lack of sufficiently skilled employees. Consequently, it is a time-consuming and inflexible process that highly depends on the domain expertise of the employees. Due to these substantial time, cost, and knowledge factors, the manual process is generally carried out irregularly and infrequently, maybe even fragmentary. This may lead to partially outdated or even obsolete prices, as current market price fluctuations are not taken into account \cite{ponnaluru2012spatial}. To reflect current market prices while supporting domain experts and digitalization of price prediction in general, it is desirable to automate the forecasting process and update the forecastings periodically.

Using \acf{ML} methods to calculate the residual value of construction equipment has already been tested in the past \cite{zong2017maintenance,chiteri2018cash,milovsevic2021estimating,Shehadeh2021,Alshboul2021}. While the results of these studies and general developments in the field of \ac{ML} are very promising, a substantial portion of the existing work originates from academic institutes, tech companies, start-ups, or large international corporations. Meanwhile, \acp{SME}, while accounting for 90\% of all businesses \cite{Ardic2011}, are not represented. Even though \acp{SME} generate large amounts of data and have significant domain knowledge, \ac{ML} applications are less common there. One of the main challenges for these organizations is the lack of skilled employees with \ac{ML} knowledge \cite{bauer2020machine}.

As an alternative to the manual creation of \ac{ML} models, \acf{AutoML} has been proposed in the last years \cite{automl_hutter2019automated,automl_yao2018taking,Zoller2021}. \ac{AutoML} aims to reduce and partially automate the necessary manual work carried out by humans when creating \ac{ML} solutions. It has already been proven to achieve good performance with a significantly smaller degree of human effort and a high computational efficiency \cite{automl_yao2018taking}. This provides a possible solution for \acp{SME} to the severe shortage of professionals with in-depth \ac{ML} knowledge.

To evaluate this potential solution, we conduct a case study in the context of used machinery valuation. Using the well-established \acf{CRISP-DM} \cite{shearer2000crisp}, we divide the different steps of creating an \ac{ML} pipeline into a non-\ac{ML} and an \ac{ML} part. The non-\ac{ML} part can be executed by domain experts, while for the \ac{ML} part, we examine different \ac{AutoML} frameworks and compare them with the traditional, manual development. The case study investigates if \ac{AutoML} is a viable alternative to manual \ac{ML} methods and how domain experts can fuel the \ac{ML} process. To easily assess our approach and create a general multimodal assessment method, we introduce the novel \acf{MES}, which incorporates different application-based metrics into one single number.

The work is structured as follows: Section~\ref{sec:related-work} presents related work. Section~\ref{sec:methodology} describes the idea of splitting the \ac{ML} pipeline into a data domain and \ac{ML} phase, introduces the manual \ac{ML} and \ac{AutoML} methods, and describes the new \ac{MES}. The main findings are presented in Section~\ref{sec:results} followed by a conclusion.

\section{\uppercase{Related work}}
\label{sec:related-work}

\subsection{Automated Price Prediction for Used Construction Machines}
Several works have been published that use \ac{ML} to calculate the residual value of construction equipment. \cite{zong2017maintenance} estimates the residual value of used articulated trucks using various regression models. Similarly, \cite{chiteri2018cash} analyses the residual value of \sfrac{3}{4} ton trucks based on historical data from auctions and resale transactions. \cite{milovsevic2021estimating} construct an ensemble model based on a diverse set of regression models to predict the residual value of \numprint{500000} construction machines advertised in the USA. \cite{Shehadeh2021} and \cite{Alshboul2021} use various regression models to predict the residual value of six different construction equipment types based on data from open-accessed auction databases and official reporting agencies.

While the results of these studies have shown first \ac{ML} successes, creating the proposed models requires \ac{ML} expertise. Our case study focuses on the potential of \ac{AutoML} and how \acp{SME} with limited \ac{ML} expertise can benefit from automated approaches in the field.

\subsection{Automated Machine Learning}
\Ac{AutoML} aims to improve the current way of building \ac{ML} applications manually via automation. While \ac{ML} experts can increase their efficiency by automating tedious tasks like \acl{HPO}, domain experts can be enabled to build \ac{ML} pipelines on their own without having to rely on a data scientist. Currently, those systems mainly focus on supervised learning tasks, \eg tabular regression \cite{Zoller2021a} or image classification \cite{Zoph2016}.

From tuning the hyperparameters of a fixed model over automatic \ac{ML} model selection up to generating complete \ac{ML} pipelines from a predefined search space, \ac{AutoML} mimics the way how humans gradually approach \ac{ML} challenges today. Virtually all \ac{AutoML} approaches formulate the automatic creation of an \ac{ML} pipeline as a black-box optimization problem that is solved iteratively \cite{Zoller2021}: potential model candidates are drawn from the underlying search space, and the performance on the given dataset is calculated. This procedure is repeated until the optimization budget, usually, a maximum optimization duration, is depleted. Often this optimization is implemented via Bayesian optimization \cite{Frazier2018}, which utilizes a probabilistic surrogate model, like a Gaussian process, to predict the performance of untested pipeline candidates and steer the optimization to better-performing regions.

\subsection{CRISP-DM}
\label{sec:crispdm-ml-pipeline}

\begin{figure*}[t]
\centering
    \includegraphics[width=\textwidth]{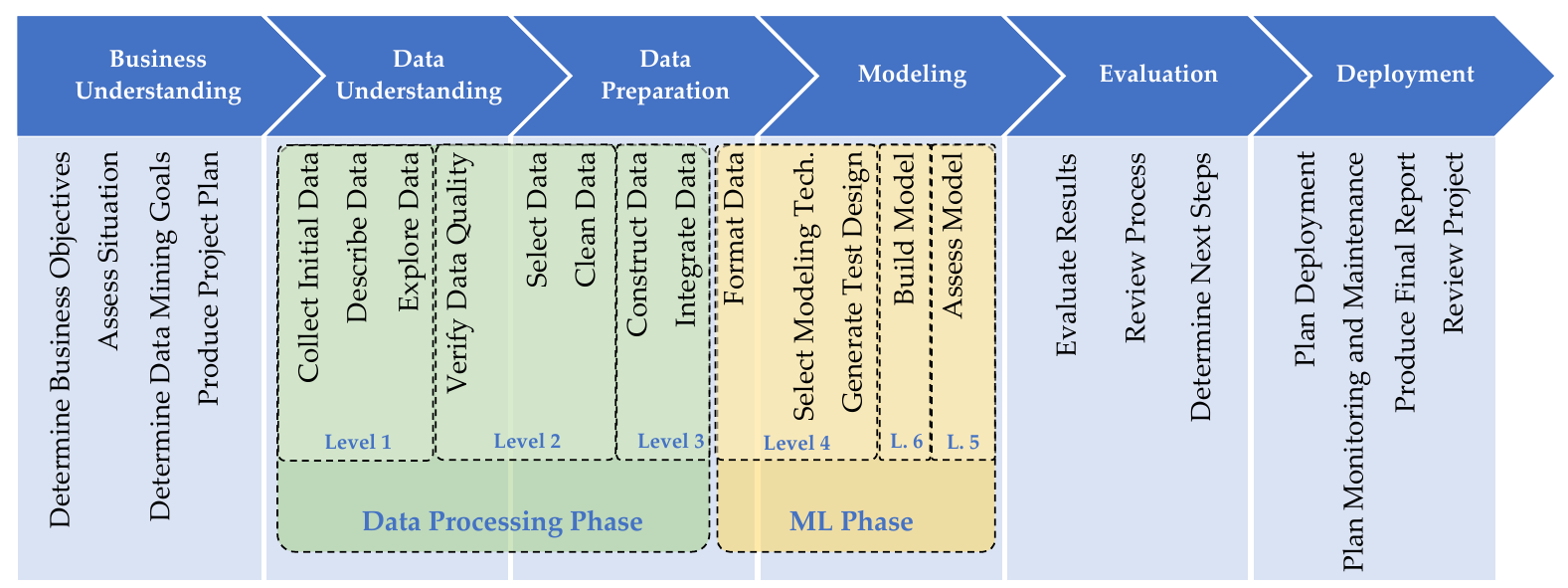}
    \caption{Adaptation of the \ac{CRISP-DM} process neglecting the loop from the original process for visual simplicity. The data understanding, preparation, and modeling steps are divided into a data processing and \ac{ML} phase; individual steps are further grouped into knowledge levels. Tasks within the data processing phase require basic programming and data processing skills and highly benefit from domain expertise. In contrast, \ac{ML} expert knowledge is mandatory for the steps within the \ac{ML} phase.}
    \label{fig:crisp-dm-process}
\end{figure*}

Due to its widespread dissemination and acceptance in data-driven development, \ac{CRISP-DM} is often used in the \ac{ML} context, for example, to develop quality assurance methodologies \cite{crispml_studer2021towards} or to investigate the interpretability of machine learning solutions \cite{crispml_kolyshkina2019interpretability}.

According to \ac{CRISP-DM}, building \ac{ML} models can be divided into six steps with associated sub-tasks, as highlighted by the blue area in Figure~\ref{fig:crisp-dm-process}. The \textit{Business Understanding} step investigates the business needs, goals, and constraints to formulate a data mining problem and develop a project plan. During the \textit{Data Understanding} phase, actual datasets are identified and collected. Also, a first analysis is conducted to understand the data and assess its quality. These datasets are further manipulated in the \textit{Data Preparation} step to properly prepare them for modeling. This includes, \eg selecting meaningful attributes, performing data cleaning, and integrating different data sources. Based on this final data set, the actual generation of the models is executed in the \textit{Modeling} step. Different models and configurations are explored, trained, and evaluated iteratively. The \textit{Evaluation} step complements the \ac{CRISP-DM} loop by verifying whether the goals defined in the business understanding step have been achieved. If the requirements are satisfied, next steps are discussed in the \textit{Deployment} phase to deploy the models to production.

\section{\uppercase{Methodology}}
\label{sec:methodology}
\subsection{Knowledge Bottleneck and Potential for Automation}
\label{subsec:bottleneck}

\ac{CRISP-DM} requires various roles---namely business analyst, data scientist, big data developer, and business owner \cite{demauro2018human}---to conduct a project. Yet, the role of a data scientist, which covers most of the data understanding, data preparation, and modeling steps of the \ac{CRISP-DM} process, is often not filled in \acp{SME}. To give a clear separation between tasks that can be performed by domain experts and tasks for which \ac{ML} expertise is needed, we divided these steps, including their associated sub-tasks, into a \textit{Data Processing Phase} and an \textit{\ac{ML} Phase}, as highlighted in Figure~\ref{fig:crisp-dm-process}. Furthermore, the tasks in the two phases can be assigned \ac{ML} knowledge levels that are required to perform them. A higher level corresponds to more required expertise. We propose the following knowledge levels for the data processing phase:

\begin{description}
    \item[Level 1] \textit{Collect, describe} and \textit{explore} data: The users can collect and store business data. They understand the data and can explore it.
    \item[Level 2] \textit{Verify, select} and \textit{clean} the data: The users understand the implications of data distributions, outliers, and missing values.
    \item[Level 3] \textit{Construct} and \textit{integrate} data: The user can construct new features out of existing data, which are potentially meaningful for the expressiveness of the dataset.
\end{description}

In contrast, the \ac{ML} phase requires knowledge levels with a profound \ac{ML} expertise:
\begin{description}
    \item[Level 4] \textit{Format data, select a model} and \textit{generate test design}: The user knows how to prepare the data for \ac{ML} methods, knows the difference between problem classes like classification and regression, and can use established \ac{ML} libraries.
    \item[Level 5] \textit{Model assessment}: The user knows the functionality and meaning of different models and can assess the impact of their hyperparameters. They must also have an understanding of the different performance metrics in regard to the dataset.
    \item[Level 6] \textit{Model creation}: The user has a deep \ac{ML} understanding, can create new models from scratch, and can optimize them via different search algorithms like grid or random search~\cite{BergstraBengio2012}.
\end{description}

After splitting the \ac{ML} pipeline into an \ac{ML} and a non-\ac{ML} phase, we analyze how \ac{AutoML} can replace the manual labor in the \ac{ML} phase. This may enable data domain experts to use \ac{ML} techniques and, consequently, speed up the development of \ac{ML} solutions within the organization significantly. \ac{AutoML} is supposed to handle as many of the more sophisticated knowledge levels as possible. Therefore, in the next sections, we explain the individual steps in more detail in the context of the residual value case study.

\subsection{Data Processing Phase}
\label{subsec:data-preparation-pahse}

\begin{table}[b]
    \caption{Collected dataset features with types and examples.}
    \label{table:features}
    \begin{tabularx}{\linewidth}{l X X}
     \toprule
     Feature            & Type          & Example \\
     \midrule
     Brand              & Categorical   & Caterpillar \\
     Model              & Categorical   & M318 \\
     Series             & Categorical   & E \\
     Construction year  & Numerical     & 2018 \\
     Working hours      & Numerical     & 8536 \\
     Location           & Categorical   & Germany \\
     Price              & Numerical     & 59.000 \texteuro \\
     \bottomrule
    \end{tabularx}
\end{table}

\paragraph{Collect, Describe \& Explore Data}
The initial data was obtained by regularly collecting all advertisements from seven major construction equipment market portals\footnote{The market portals are \href{https://www.mascus.de/}{Mascus}, \href{https://catused.cat.com/}{Catused}, \href{https://www.mobile.de/}{Mobile}, \href{https://machineryline.de/}{MachineryLine}, \href{https://trademachines.de/}{TradeMachines}, \href{https://www.truck1.eu/}{Truck1}, and \href{https://www.truckscout24.de/}{Truckscout24}.} over a time period of seven months. In total, \numprint{11606162} entries from different manufacturers have been collected. The collected features, selected by data domain experts a priori, are shown in Table~\ref{table:features}.

\begin{figure}[t]
\centering
    \includegraphics[width=\linewidth]{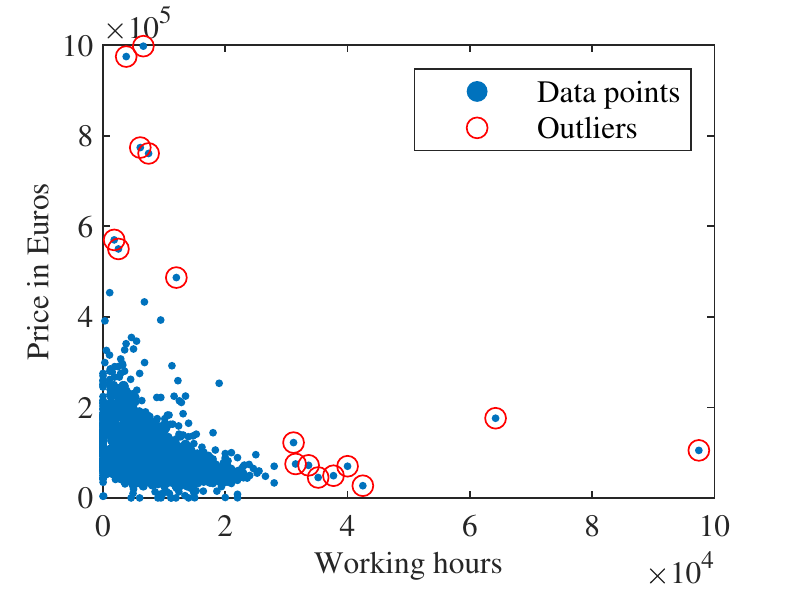}
    \caption{Working hours versus price where detected outliers are highlighted.}
    \label{fig:data-scatter-plots}
\end{figure}

\paragraph{Verify Data Quality}
A drawback of collecting data automatically by web-scraping is the resulting dataset quality. Regular collection of advertisements from web portals leads to duplicated data points, as the same construction machine can be offered on different platforms and for longer periods. Furthermore, the quality and completeness of the dataset depend on the input of the portal users, which may lead to incorrect or missing attributes. Outliers were primarily present in the attributes \textit{working hours} and \textit{price}, as displayed in Figure~\ref{fig:data-scatter-plots}.

\paragraph{Select \& Clean Data}
Duplicate entries are eliminated by an iterative comparison of different feature combinations. In the \textit{Working hours} feature, outliers were identified through reviews by the respective domain experts, considering the average number of operating hours for the given model and year of manufacture. For instance, some machines were advertised with operating hours much larger than their expected lifetime. For outliers regarding the \textit{price} feature, the main source of noise was traced back to a missing currency conversion for one of the biggest dealers from Poland. These outliers were detected by a plausibility check---namely, removing values outside a $99$\% confidence interval---considering the working hours and price. Errors in the \textit{series} attribute are also mainly caused by incorrect inputs by the selling dealer. If the underlying reason for errors or outliers could not be determined, the sample was dropped.

Dealing with missing values depends on the attribute. Samples are dropped if a value of the features \textit{model}, \textit{construction year}, or \textit{location} is missing. Missing values for the \textit{working hours} attribute will be substituted via stochastic regression imputation \cite{newman2014missing}. Values for the \textit{series} attribute are optional. The entries for \textit{brand} and \textit{price} are mandatory on all portals for creating advertisements.

\begin{table*}[ht]
    \caption{Excerpt of the final dataset.}
    \label{table:data-excerpt}
    \begin{tabularx}{\textwidth}{X X X l l X X}
    \toprule
    Brand       & Model & Series    & Construction~year & Working~hours & Location & Price\\
    \midrule
    Caterpillar & 308   & D         & 2010 & 4865   & BE & 38500 \\
    Caterpillar & D6    & T         & 2016 & 11851  & NL & 112000 \\
    Caterpillar & 330   & F         & 2015 & 4741   & CH & 126056 \\
    Caterpillar & M318  & F         & 2016 & 8920   & PL & 99000 \\
    Caterpillar & 966   & K         & 2012 & 10137  & FR & 82000 \\
    \bottomrule
    \end{tabularx}
\end{table*}

\paragraph{Construct \& Integrate Data}
Finally, to ensure sufficient data for each construction machine model, only model types with more than \(150\) samples were added to the dataset, resulting in \(10\) different machine models and \(2910\) samples in total.
As all remaining machine models are manufactured by Caterpillar, the \textit{brand} feature, depicted in Table \ref{table:features} is thus obsolete. An excerpt of the resulting dataset is shown in Table~\ref{table:data-excerpt}. To also account for and investigate the impact of single features, data subsets with individual feature combinations are created. The subset consisting of the machine model, working hours, and construction year is used as the baseline feature set (subsequently referred to as \textit{basic subset}). In addition, this basic subset was extended by the series and/or location feature, resulting in four data sets.

\subsection{ML Phase}
\label{subsec:ml-phase}
\begin{figure}[t]
\centering
    \includegraphics[width=\linewidth]{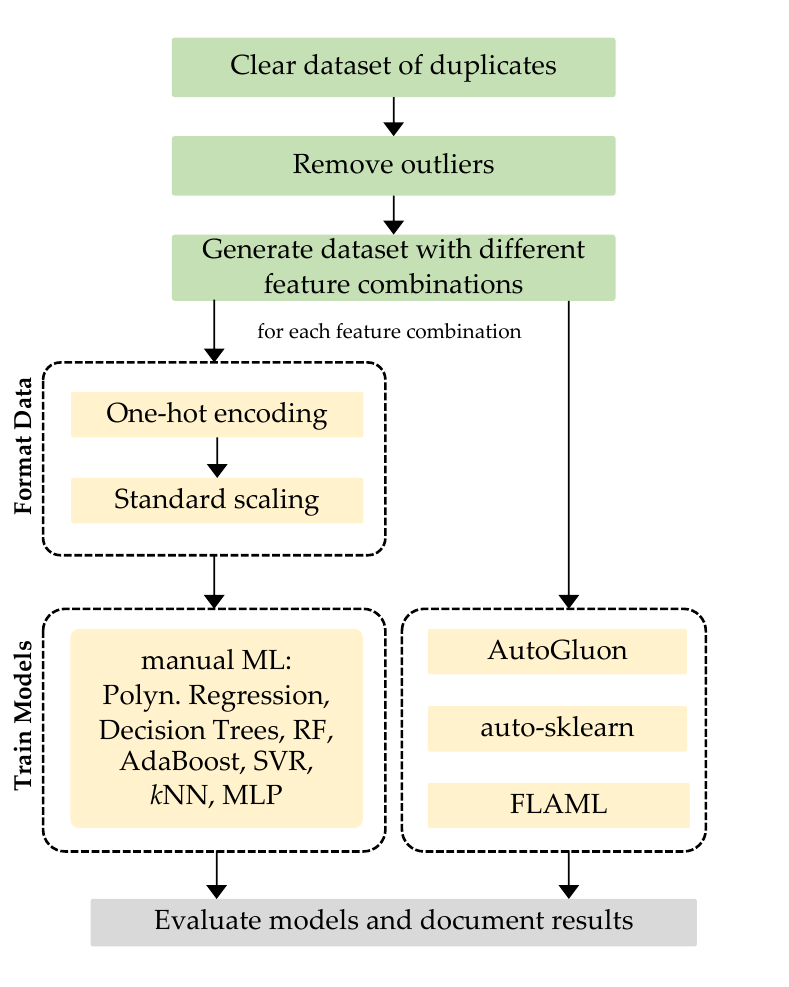}
    \caption{The case study pipeline illustrates the steps of the data processing phase in green and the \ac{ML} phase in yellow. One-hot encoding and standard scaling are only applied to the manually implemented \ac{ML} methods.}
    \label{fig:framework}
\end{figure}

As we want to test whether \ac{AutoML} can be a potential substitution for manual \ac{ML}, we describe the manually implemented pipeline and selected \ac{AutoML} frameworks in more detail below. An overview of the whole \ac{ML} pipeline for the case study is depicted in Figure~\ref{fig:framework} and the source code is available on Github\footnote{
    See \url{https://tinyurl.com/4wt2hp2y}.
}.

\subsubsection{Manual ML}
The manually created pipeline closely resembles a best-practice pipeline \cite{geron2022hands}. All features are pre-processed using \textit{one-hot encoding} for categorical features and \textit{standard scaling} for numerical features. Next, seven different state-of-the-art and well-established \ac{ML} methods are used for predicting the residual values:
\begin{itemize}
    \item Polynomial Regression
    \item Tree-based regression: decision tree, \ac{RF} \& adaptive boosting (AdaBoost)
    \item Kernel-based regression: \ac{SVR} \& \ac{kNN}
    \item Deep learning: \ac{MLP}
\end{itemize}
For each of those methods, selected hyperparameters are tuned via random search \footnote{
The list of hyperparameters is available within the online source code.
}. For more information on those \ac{ML} methods, we refer the interested reader to \cite{geron2022hands}.

\subsubsection{Automated ML}
The manual approach is compared with the three open-source \ac{AutoML} frameworks \name{AutoGluon} \cite{agtabular}, \name{auto-sklearn} \cite{feurer2020auto2}, and \name{FLAML} \cite{wang2022automated}. We chose those \ac{AutoML} frameworks as they are
\begin{enumerate*}
    \item simple to use,
    \item well documented,
    \item easy to integrate,
    \item have achieved good performances in the past \cite{Gijsbers2019}, and
    \item have a broad user base and, therefore, good support.
\end{enumerate*}
To ensure a fair comparison, we restricted the selection of algorithms to the ones with implementation in the same underlying \ac{ML} library, namely scikit-learn \cite{Pedregosa2011}, and omitted, for example, \ac{NNI} \cite{Microsoft_Neural_Network_Intelligence_2021} as it uses additional frameworks, \eg PyTorch \cite{Paszke2019}.

The selected frameworks promise an end-to-end creation of \ac{ML} pipelines, including all necessary pre-processing steps, for tabular regression tasks. Consequently, data is not manually pre-processed.

\subsection{Criteria}
\label{subsec:metric}

To determine the relative performance of the models, we define a novel benchmarking scheme. In literature and practice, a multitude of commonly used evaluation criteria---such as \ac{MAE}, \ac{RMSE}, and \ac{MAPE}---are well known and widely adopted to assess the performance of an \ac{ML} regressor. To account for industrial requirements, these technical metrics have to be complemented by further non-technical ones. Consequently, multiple factors have to be integrated into the algorithm selection process. Following \cite{Ali2017}, we define quality metrics with application-based meanings that domain experts can understand:

\begin{description}
    \item[Correctness] ($s_{\text corr}$) measures the predictive power of an \ac{ML} model. This corresponds to typical metrics used in supervised learning. In the context of this work, the \ac{MAPE}
    \begin{equation*}
        \text{MAPE} = \frac{1}{n}\sum_{i=1}^{n}\frac{|y_{i} - \hat{y_{i}}|}{|y_{i}|}
    \end{equation*}
    is used to calculate the performance, with \(y_{i}\) being the true value, \(\hat{y_{i}}\) the predicted value and \(n\) the number of samples.
    \item[Complexity] ($s_{\text comp}$) measures the training complexity of an \ac{ML} model. In this work, we use the CPU wall-clock time as a proxy for this metric.
    \item[Responsiveness] ($s_{\text resp}$) measures the inference time of an \ac{ML} model by determining the CPU wall-clock time required to create a single prediction. This aspect may be especially important for interactive and real-time systems. Following \cite{Nielsen1993}, runtimes are mapped into a real-time (under 0.1 seconds), fast (under 1 second), and slow (above 1 second) category.
    \item[Expertise] ($s_{\text exp}$) measures the knowledge level, as introduced in Section~\ref{subsec:bottleneck}, required to be able to create the according \ac{ML} solutions in the first place.
    \item[Reproducibility] ($s_{\text repr}$) measures the stability of the \ac{ML} model regarding the other criteria by determining the standard deviation if retrained on the exact same data again.
\end{description}

These criteria are combined into a novel score to create a ranked list of \ac{ML} models. It is, therefore, mandatory to normalize all criteria values to \([0, 1]\) using min-max scaling. Furthermore, the actual metrics in each criterion have to be compatible with each other by having identical optimization directions; in our case smaller values being better. Preferences regarding the weighting of individual criteria should be incorporated into the final score. This can be done by assigning weights \(w_c\) to each criterion leading to the final \textit{\acf{MES}} using the weighted average
\begin{equation}
\label{equ:mes}
    \text{MES} = \dfrac{\sum_{c \in C} w_c \: \Tilde{s_c}}{\sum_{c \in C} w_c}
\end{equation}
with \(C = \{\text{corr} \mathrm{,\:} \text{comp} \mathrm{,\:} \text{resp} \mathrm{,\:} \text{exp} \mathrm{,\:} \text{repr}\}\) and $\Tilde{s_c}$ being the normalized values. By design, the \ac{MES} is bound to \([0, 1]\), where zero indicates a perfect and one the worst performance with regard to all individual metrics. To ensure reliable results and make the calculation of reproducibility even possible, models need to be fitted multiple times.

\section{\uppercase{Results}}
\label{sec:results}
This section presents the results of the experiments. For a better overview, we only present the two best (in terms of correctness) \ac{ML} models, \ac{RF} and \ac{MLP}, out of the seven examined manual methods and use them as a baseline for comparison against the investigated \ac{AutoML} frameworks. All measurements were performed on a Ubuntu Linux 20.04.5 LTS system with $32$ GB RAM and an Intel i7-4790 Processor. We conducted five independent measurements with fixed $90$\% / $10$\% holdout training/test split.

\begin{figure*}[t]
\centering
    \includegraphics[width=\textwidth]{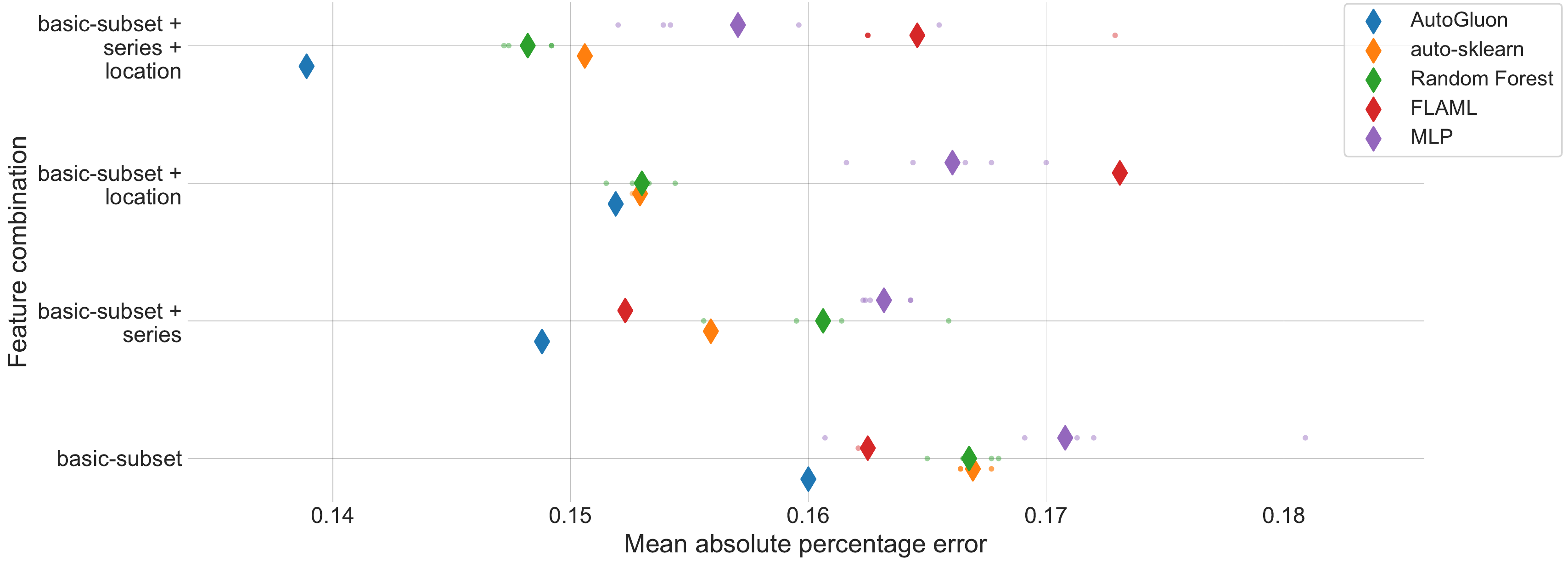}
    \caption{Correctness, in form of \acs{MAPE}. Diamonds depict the average results over $5$ repetitions. Single measurements are displayed as dots. Due to low variance, single measurements are often not visible. For better visibility, only the winning \acs{RF} method is shown for the set of the manual \ac{ML} methods.}
    \label{fig:mape-scatter-plot}
\end{figure*}

\begin{table*}[bp]
    \caption{\ac{MES} with unnormalized underlying criteria for all methods on the complete feature set. Smaller values are better. The best results are highlighted in bold. The observed differences in the results for each criterion were significant according to a Student's \(t\)-test with \(\alpha = 0.05\). \name{Reproducability}, defined as the standard deviation of \name{Correctness}, is not depicted as its own column. Results marked by * did produce constant results.}
    \label{table:result-metric-results}
    \begin{tabular*}{\textwidth}{@{\extracolsep{\fill}}l c c c c c c c}
     \toprule
     Method                     & Correctness [MAPE]             & Complexity [sec.]         & Expertise     & Responsiveness   & \ac{MES} \\
     \midrule
     \ac{MLP}                   & ~~~~~~\res{0.1570}{0.0049}     & ~~~~\res{2308.1}{354.70}  & 5             &\textbf{real-time}     & ~\res{0.977}{0.0130} \\
     \ac{RF}                    & ~~~~~~\res{0.1482}{0.0009}     & ~~\res{1067.6}{29.93}     & 5             &\textbf{real-time}     & ~\res{0.896}{0.0138} \\
     \name{auto-sklearn}        & 0.1506 *~~~~~~                 & \res{1806.1}{3.52}        & \textbf{2}    &\textbf{real-time}     & ~\res{0.696}{0.0160} \\
     \name{AutoGluon}           & \textbf{0.1389} *~~~~~~        & ~~~~\bres{14.2}{0.17}     & \textbf{2}    &\textbf{real-time}     & ~\bres{0.583}{0.0101} \\
     \name{FLAML}               & ~~~~~~\res{0.1646}{0.0042}     & \res{1801.5}{0.71}        & \textbf{2}    &\textbf{real-time}     & ~\res{0.738}{0.0108} \\
     \bottomrule
    \end{tabular*}
\end{table*}

\subsection{Correctness}
\label{subsec:results-prediction-quality}

The correctness, in form of the \ac{MAPE}, of the different approaches is depicted in Figure~\ref{fig:mape-scatter-plot}. \name{AutoGluon} delivers the best results for all feature combinations. For the rest of the methods, there is no clear trend or order. Thus, concerning the prediction quality, the \ac{AutoML} methods are comparable to or even better than the manual \ac{ML} methods. The best results with respect to minimal predictive error for all methods are achieved with the entire feature set.

\subsection{Expertise}
\label{subsec:results-implementation-complexity}

Implementing and tuning the seven manual \ac{ML} methods presumes expertise of level $5$ and requires approximately 50 \ac{LOC} on average. The manual approaches must be implemented and configured by hand and require a profound understanding of the different ML libraries, their functionalities, and when and how to use them. On the other hand, training and predicting using \name{AutoGluon}, \name{auto-sklearn}, or \name{FLAML} can be implemented within $5$ \ac{LOC} and without any \ac{ML} expertise. This demonstrates that basic programming knowledge is sufficient to use the \ac{AutoML} frameworks. Yet, generating and storing the data is still necessary (knowledge level $1$). The same holds true for verifying and cleaning the data (knowledge level $2$). Thus, the knowledge demands for the \ac{AutoML} methods, with level $2$, and the manual \ac{ML} methods, with level $5$, are quite different, with a clear advantage for the \ac{AutoML} methods.

\subsection{Responsiveness}
Responsiveness is measured as the average prediction time over all samples. Predictions of a single sample are always in a millisecond range. Consequently, all methods fall in the real-time application category.

\subsection{Complexity}
\label{subsec:results-runtime}

\begin{figure*}[t]
\centering
    \includegraphics[width=\textwidth]{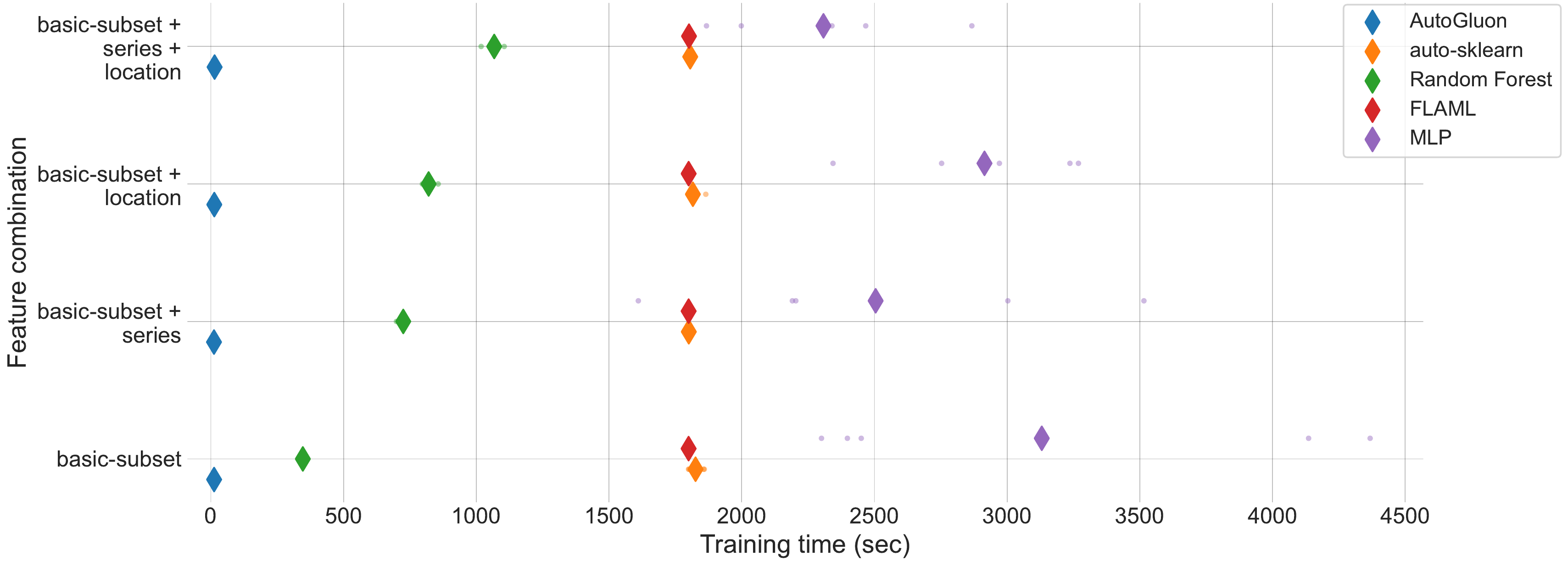}
    \caption{Model complexity, in terms of training duration, for all feature set combinations. The \ac{AutoML} frameworks are configured to optimize for 1800 seconds. Diamonds depict average results over $5$ repetitions. Single measurements are displayed as dots. Due to low variance, single measurements are often not visible. For better visibility, only the winning \ac{RF} method is shown for the set of the manual \ac{ML} methods.}
    \label{fig:runtime-results}
\end{figure*}

The results for the method complexity, in terms of training duration, are depicted in Figure~\ref{fig:runtime-results}. \name{AutoGluon} has the lowest training time with about $15$ seconds, being much better than \ac{RF} coming in second place. In contrast to all other analyzed methods, \name{AutoGluon} does not search for an optimized model but trains only a single predefined ensemble. \name{FLAML} and \name{auto-sklearn} fully utilize the specified training budget of $1800$ seconds, whereas the manual \ac{ML} methods are controlled by an iteration number and not by a time limit. The detailed algorithmic analysis of these findings is further analyzed in a follow-up work.

\subsection{Reproducibility}
As the values of the expertise (\(s_{\mathrm{exp}}\)) and responsiveness (\(s_{\mathrm{resp}}\)) are categorical measurements, we did not observe any variance making these criteria unsuited. Both correctness (\(s_{\mathrm{corr}}\)) and complexity (\(s_{\mathrm{comp}}\)) expressed usable variance. In the context of this work, we decided to use correctness as the basis for reproducibility (\(s_{\mathrm{repr}}\)). A variance of performance can be observed for \name{FLAML}, \ac{RF}, and \ac{MLP}, while both \name{auto-sklearn} and \name{AutoGluon} produced constant results.

\begin{figure}[b]
\centering
    \includegraphics[width=0.98\linewidth]{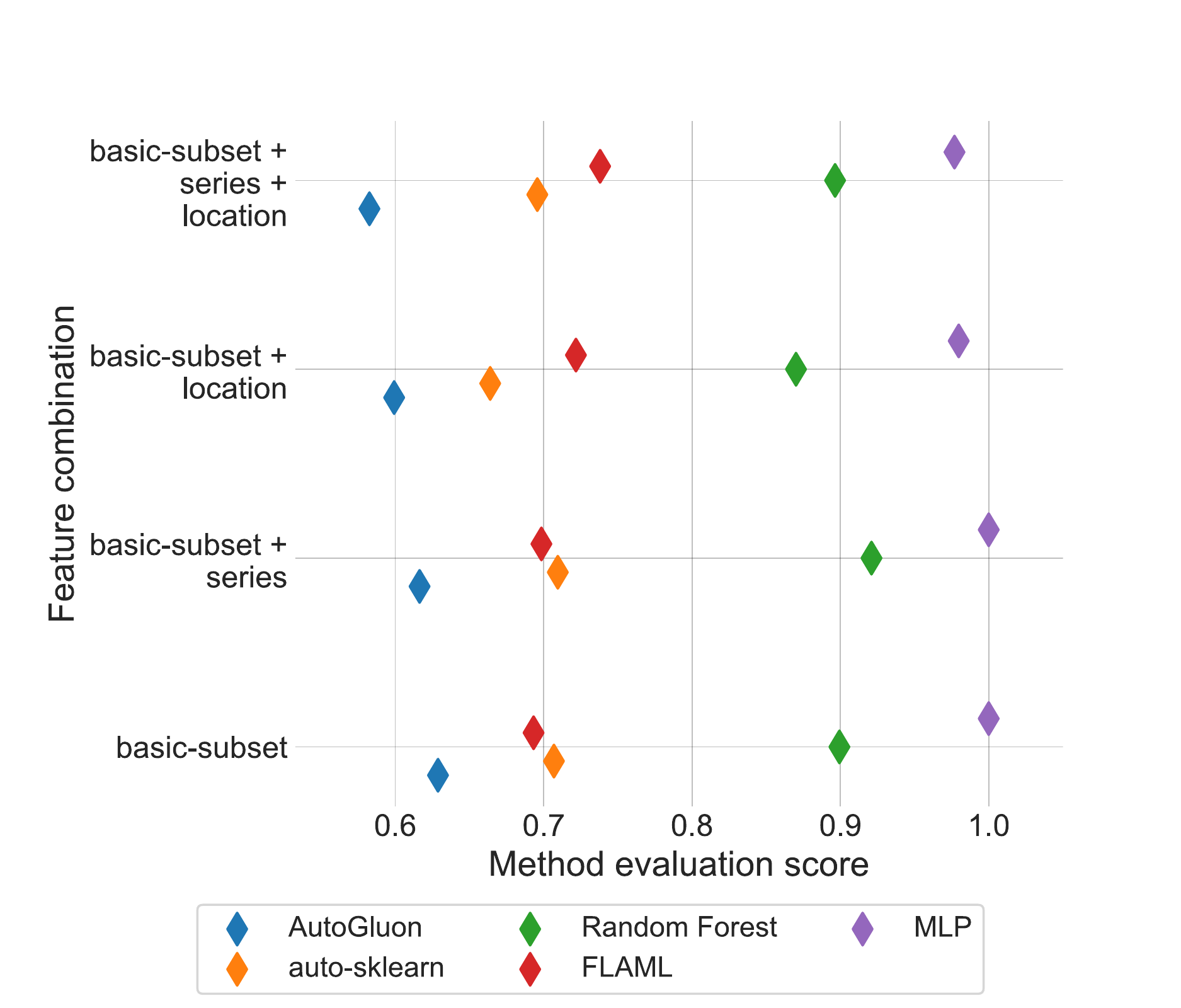}
    \caption{\Acl{MES} for each method and subset combination. Smaller values are better.}
    \label{fig:mes-scatter-plot}
\end{figure}

\subsection{Method Evaluation Score}
\label{subsec:result-method-evaluation}
We determined the values of the weighting factors for individual criteria by surveying six domain experts, averaging the results, and rounding them to the nearest tens for simplicity. In our case, $w_{\mathrm{corr}} = 50$, $w_{\mathrm{exp}} = 40$ and $w_{\mathrm{comp}} = 10$. While in general, all criteria are important, $w_{\mathrm{resp}}$ and $w_{\mathrm{repr}}$ are set to $0$ for the considered use case because they were deemed unimportant by the domain experts. The \acp{MES}, as defined in Equation~\eqref{equ:mes}, are calculated for each method/feature combination and depicted in Figure~\ref{fig:mes-scatter-plot}. For the feature subset with the lowest overall \ac{MES}, the individual criteria scores and the final \acp{MES} are depicted in Table~\ref{table:result-metric-results}. The winning method/subset combination is \name{AutoGluon} and the complete subset, with an \ac{MES} of $0.583$. The \ac{MES} drastically simplifies the methods' comparability and shows that \name{AutoGluon} is performing best for the given data set and weighting factors. Based on these findings, \ac{AutoML} seems to be a good alternative for this use case.

\section{\uppercase{Conclusion}}
\label{sec:conclusion}

This work analyzed the potential of \ac{AutoML} methods and their usability for \acp{SME} with limited \ac{ML} expertise. In our case study, predicting the residual value of used heavy construction equipment, all evaluated \ac{AutoML} methods were shown to outperform manually created \ac{ML} pipelines regarding the newly introduced \acf{MES}, see Equation~\eqref{equ:mes}. Furthermore, they are applicable with only domain knowledge and basic data processing skills. We, therefore, showed that separating the data understanding, data preparation, and modeling steps of the \ac{CRISP-DM} process into a data domain and an \ac{ML} part enables companies with limited \ac{ML} expertise to tackle \ac{ML} projects by using \ac{AutoML} methods. We introduced \ac{ML} expertise levels and used the \ac{MES} to enable an easy assessment of the different \ac{ML} and \ac{AutoML} methods.

To transfer the results identified in this case study to other use cases, a qualification of domain experts for at least knowledge level $2$ is necessary. In summary, the evaluation of the models created by \name{AutoGluon} was deemed favorable. The predictions were validated by the domain experts as valid and reliable. Consequently, the deployment phase in the \ac{CRISP-DM} process can be planned and implemented.

It has to be mentioned that we only examined a limited number of \ac{ML} and \ac{AutoML} methods on four variations of a single data set, so that general statements are therefore limited by our choice of methods. In the evaluated use case \ac{AutoML} was able to provide results with a good performance, yet it still may not be applicable for some use cases. 
\ac{AutoML} tools may create models with low predictive power or even fail to generate a model at all. To resolve some of these issues, knowledge of \ac{ML} could be necessary, which users with knowledge level three or lower do not have.

In the future, we plan to examine the differences between the \ac{AutoML} methods in more detail and extend their usability for \acp{SME} by adding additional preprocessing steps like data splitting. In addition to the \ac{MES}, we aim to develop a data-centric explanation of the final results to provide more insights for domain experts. This is intended to explain the model behavior via the dataset and should enable the domain experts to validate the quality and reliability of the results based on the data used to train the models. These data-centric explanations are crucial in order to generate confidence in the results and increase the willingness of domain experts to use \ac{AutoML} methods.

\section*{ACKNOWLEDGEMENTS} 
This work was partly funded by the German Federal Ministry of Economic Affairs and Climate Action in the research project \name{AutoQML}.

\bibliographystyle{apalike}
{\small
\bibliography{library_horst}}

\end{document}